\documentclass[10pt, a4paper]{article}

\usepackage[final]{lrec2026} 
\usepackage{tabularx}
\usepackage{multirow}
\usepackage{booktabs}
\usepackage{soul}
\usepackage[safe]{tipa}
\usepackage{verbatim}
\usepackage[T1]{fontenc}
\usepackage[utf8]{inputenc}
\usepackage{float}
\title{From Rosetta to Match‑Up: A Paired Corpus of Linguistic Puzzles with Human and LLM Benchmarks}

\name{Neh Majmudar\textsuperscript{1}, Anne Huang\textsuperscript{2}, Jinfan Frank Hu\textsuperscript{3}, Elena Filatova\textsuperscript{1}} 

\address{\textsuperscript{1}City University of New York (CUNY), \textsuperscript{2}Davidson Academy, \textsuperscript{3}Phillips Academy \\
         nmajmudar@gradcenter.cuny.edu\\
         }

\abstract{
In this paper, we examine linguistic puzzles used in high school linguistics competitions, focusing on two common formats: \textit{Rosetta Stone} and \textit{Match-Up}. We propose a systematic procedure for converting existing Rosetta Stone puzzles into corresponding Match-Up counterparts. Because linguistic puzzle creation is complex and time-consuming, our method provides an efficient way to accelerate the generation of new puzzles. We evaluate the resulting Rosetta Stone–Match-Up pairs with both human participants and large language models (LLMs). Our results show that both expert human solvers and LLMs display an all-or-nothing pattern on Match-Up puzzles, either solving them completely or failing entirely. This work contributes a new dataset of paired puzzles and provides a detailed evaluation of puzzle difficulty across formats, offering insights into both human and machine linguistic reasoning.
 \\ \newline  }

\begin{document}

\maketitleabstract

\section{Introduction}
\label{sec:Intro}

\begin{table*}[t!]
\centering
\begin{tabularx}{\textwidth}{lll} 
 \toprule
\multicolumn{3}{p{6in}}{The Gilbertese puzzle used in UKLO in 2018. This puzzle has two difficulty scores: its score for the Breakthrough participants is $34\%$ and its score for the Foundation participants $59\%$; its linguistic topic is syntax; its language family is Austronesian, Oceanic; its Author is Michael Salter.}\\
 \hline
 \multicolumn{1}{c}{\textbf{ }} & 
\multicolumn{1}{c}{\textbf{Gilbertese}} &
\multicolumn{1}{c}{\textbf{English}} \\
 \hline
1.  & \textbf{Ko nakonako \ng{}koe} & \textit{You are walking} \\ 
 \hline
2.  & \textbf{E nakonako te aiine}   & \textit{A woman is walking} \\ 
 \hline
3.  & \textbf{I takaakaro \ng{}ai }  & \textit{I am playing} \\ 
 \hline
4.  & \textbf{E nakonako nakon te titooa Meeri}   & \textit{Mary is walking to the store} \\ 
 \hline
5.  & \textbf{A tekateka irarikin te auti aiine}  & \textit{Women are sitting next to the house} \\ 
 \hline
6.  & \textbf{A tebotebo nakekei n te bong aei}   & \textit{People are bathing today} \\
 \hline
7.  & \textbf{I tebotebo inanon te auti \ng{}ai}  & \textit{I am bathing in the house} \\ 
 \hline
8.  & \textbf{A takaakaro inanon te titooa ataei}   & \textit{Children are playing in the store} \\ 
 \hline
9.  & \textbf{Ko tekateka \ng{}koe ningaabong}   & \textit{You will sit tomorrow} \\ 
 \hline

10.  & \textbf{E takaakaro irarikin te kawai te ataei }   & \textit{The child is playing next to the road today} \\ 
      & \textbf{n te bong aei}   & \textit{ } \\ 
 \hline

\textbf{Q.5.3} & \underline{Translate from English into Gilbertese:}  & \\

11. &  \textit{Women will play tomorrow.}  & \\
12. &  \textit{You are sitting next to the store today.} &  \\
\bottomrule
\end{tabularx}
\caption{Rosetta Stone Linguistic Puzzle Example}
\label{tab:RosettaStoneExample}
\end{table*}

Linguistic puzzles designed for high school–level competitions, such as the International Linguistics Olympiad (IOL)\footnote{\url{https://ioling.org/}} and various national contests, are now used not only to assess the skills of high school students and other linguistics enthusiasts but also as benchmarks for evaluating the performance of Large Language Models (LLMs)~\cite{LingOly}. Thus, studying these puzzles serves a dual purpose: advancing the popularization of linguistics and providing a testbed for both the technical capabilities and the creative potential of LLMs.

One persistent challenge for both human solvers and LLMs is the relatively limited supply of existing puzzles. Creating high-quality puzzles is a creative and engaging process but also labor-intensive, often requiring the expertise of highly skilled linguists to ensure validity. This difficulty is further compounded by the absence of formal, widely accepted criteria for evaluating puzzle quality~\cite{Gleason,Zaliznyak,WordLetterNumber,bozhanov-derzhanski-2013-rosetta}.

To ensure fairness for all competition participants, puzzles created for both national and international linguistics competitions are typically based on languages unlikely to be familiar to them. The puzzle problem statements are written in the national language of the host country, and participants are not expected to know any foreign languages. In this project, we focus on puzzles prepared for English-speaking participants. 

Because generating complex and engaging linguistic puzzles requires a high degree of creativity, the puzzle generation task cannot be fully formalized. This limitation places competition, level puzzle generation beyond the current capabilities of even the most advanced LLMs. Each linguistic puzzle can be characterized along several dimensions, including its difficulty, central linguistic topic, language, and format. In this work, we focus on the last of these—puzzle format—and seek to answer the following question: does each format require its own dedicated generation procedure, or can a single approach accommodate multiple formats?

There are several established formats of linguistic puzzles, including Rosetta Stone, Match-Up (also known as Chaos), Monolingual, Pattern, Computational, and Text. The definitions below are taken \textit{verbatim} from the UKLO website.\footnote{\url{https://www.uklo.org/technical-information/\#qformat}}
\begin{itemize}
\item \textbf{Rosetta}: The data are sets of corresponding words or phrases between different languages/writing systems, with most of the correspondences given. Parts may be omitted from the data set, leaving gaps to be filled. You must be able to give new correspondences (typically translations) to solve the task.
\item \textbf{Match-Up}: The data are sets of corresponding words or phrases in multiple languages/writing systems, but with few of the correspondences given. If some words are not part of a set, it can still count as a match-up. You must be able to give new correspondences (typically translations) to solve the task.
\item \textbf{Monolingual}: The data are texts in an unknown language (or equivalent), with no direct translation ( or transliteration for writing systems) given, with the possible exception of 1-2 words. You must be able to translate from the language to solve the task.
\item \textbf{Pattern}: The data are words or sets of forms of words/cognates, conforming to a pattern (possibly with some exceptions). You must be able to give other words conforming to this pattern or identify outliers to solve the task, but (unlike in a Rosetta) there is no translation component.
\item \textbf{Computational}: The problem data includes a description of a computational or other logical system. To solve the problem, you must be able to analyse and implement this system.
\item \textbf{Text}: The data are whole texts presented in different languages or scripts, but not subdivided further. To solve the problem, you must use context and other clues to deduce linguistic rules.
\end{itemize}

\begin{table*}[t!]
\centering
\begin{tabularx}{\textwidth}{llll} 
 \toprule
\multicolumn{4}{p{6in}}{The Polish puzzle used in UKLO in 2015. This puzzle has two difficulty scores: its score for the Breakthrough participants is $58\%$ and its score for the Foundation participants $75\%$; its linguistic topic is syntax; its language family is Indo-European, Balto-Slavic; its Author is Daniel Rucki. }\\
 \hline
 \multicolumn{1}{c}{\textbf{ }} & 
\multicolumn{1}{c}{\textbf{Polish}} &
\multicolumn{1}{c}{ } &
\multicolumn{1}{c}{\textbf{English}} \\
 
 \hline
A & \textbf{Alicja zobaczy\l{}a s\k{a}siada.} &  1 & \textit{The cat saw the mouse.} \\ 
 \hline
B & \textbf{Kot zjad\l{} kie\l{}bas\c{e}.} &  2 & \textit{Peter bought the sausage.} \\ 
 \hline
C & \textbf{Piotr kupi\l{} kie\l{}bas\c{e}. } & 3 & \textit{Alice bought the cheese.} \\ 
 \hline
D & \textbf{Mysz zobaczy\l{}a s\k{a}siada. } &  4 & \textit{Alice saw the neighbour.} \\ 
 \hline
E & \textbf{Kot zobaczy\l{} mysz. } &  5 & \textit{The mouse saw the neighbour. } \\ 
 \hline
F & \textbf{Alicja kupi\l{}a ser. } &  6 & \textit{The cat ate the sausage. } \\
  \hline
\multicolumn{4}{p{6in}}{\textbf{Q.3.1.} Pair each Polish sentence with its English translation in the table below; for example, if you think Polish sentence A is translated by English sentence 1, write `1' in the box below A. }\\
\\
\multicolumn{4}{c}{
\begin{tabular}{|l|l|l|l|l|l|l|}
        \hline
        Polish & A & B & C & D & E & F \\
        \hline
        English &  &   &    &   &   &  \\
        \hline
\end{tabular}
} \\
\bottomrule
\end{tabularx}
\caption{Match-Up Linguistic Puzzle Example}
\label{tab:MatchUpExample}
\end{table*}

Among linguistic puzzle types, Rosetta Stone and Match-Up are the most common~\cite{bozhanov-derzhanski-2013-rosetta}. Table~\ref{tab:RosettaStoneExample} presents an example of a Rosetta Stone puzzle, while Table~\ref{tab:MatchUpExample} shows a Match-Up puzzle. Both puzzles originally included additional questions; however, for clarity of illustration, we retain here only the parts corresponding to the Rosetta Stone and Match-Up format respectively. The original Rosetta Stone was used in UKLO in 2018.\footnote{\url{https://www.uklo.org/wp-content/uploads/2022/05/2018_5-Gilbertese.pdf}} The original Match-Up puzzles was used in UKLO in 2015.\footnote{\url{ https://www.uklo.org/wp-content/uploads/2022/05/2015_3.-Polish.pdf}}

National linguistics competitions, including the United Kingdom Linguistics Olympiad (UKLO) and the North American Computational Linguistics Open Competition (NACLO\footnote{\url{https://naclo.org/}}), are explicitly designed so that no prior knowledge of linguistics or specific foreign languages is required. The UKLO website states that its questions require no prior linguistic training, and archival problem sets demonstrate that puzzles frequently draw on languages that are unfamiliar to most participants, such as Beja, Lezgian, Fur, Saisiyat, and Kavalan. In addition, several UKLO problems are based on constructed languages, including Afrihili, Blazon, Esperanto, Centauri, and Arcutan. Some of these (e.g., Centauri and Arcutan) were created specifically for individual competition problems, while others (e.g., Esperanto and Afrihili) are historically documented attempts to develop regularized international auxiliary languages.

More broadly, linguistic olympiad competitions emphasize the use of low-resource, typologically diverse, or otherwise unfamiliar languages in order to evaluate analytical reasoning rather than memorized linguistic knowledge. This design principle makes linguistics puzzles a particularly valuable benchmark for evaluating large language models (LLMs), which may otherwise rely on parametric knowledge of widely documented languages instead of performing genuine pattern induction from the data provided.

A quantitative survey of UKLO problem sets further illustrates this design choice. Among 235 problems published on the UKLO website between 2010 and 2025, we identified 206 unique languages. Of these, 186 languages appear in only a single puzzle, 16 languages appear in two puzzles, and only 3 languages appear in three puzzles. Modern English is explicitly specified as the primary object of analysis in 8 puzzles; however, only one of these is a Rosetta Stone–style problem, involving the encoding of English using a four-digit cipher. The overwhelming predominance of one-off language usage underscores the deliberate avoidance of repetition and prior familiarity, reinforcing the competition’s emphasis on in-problem reasoning rather than accumulated language knowledge.

The UKLO puzzles are fairly balanced across the major linguistic domains of phonology, semantics, morphology, syntax, and writing systems. Computational puzzles and those involving number systems are less frequent. It should be noted that many puzzles span multiple linguistic topics rather than fitting neatly into a single category. 

%


In this work, we investigate if the Rosetta Stone and Match-Up formats represent fundamentally the same type of linguistic puzzles viewed from two different perspectives. Answering this question can inform whether a single generation procedure could suffice for both of these formats or whether separate, format-specific procedures are necessary.

For our study, we use a collection of Rosetta Stone puzzles along with their solutions that are listed on the UKLO website. For each puzzle, we take the original problem statement, the associated questions, and the provided answers, and transform them into a corresponding Match-Up puzzle. One outcome of this work is a curated corpus of Rosetta Stone puzzles paired with their corresponding Match-Up versions.

We then select two subsets of the Rosetta Stone, Match-Up puzzle pairs and ask two high school students, both experienced linguistic puzzle solvers, to solve the puzzles from these subsets. Our results show no performance difference (drop) for the solvers when working with original Rosetta Stone versus synthetic Match-Up puzzles. We subsequently provide the complete set of Rosetta Stone/Match-Up pairs to LLMs for evaluation.

We also document the strategies the human solvers use when solving puzzles. We see three key benefits to recording these strategies:
\begin{enumerate}
    \item They provide insight into what constitutes a well-designed linguistic puzzle.
    \item They can inform and guide the development of future puzzle-generation procedures.
    \item They can support the analysis of LLMs’ chain-of-thought reasoning, helping to better understand differences in decision-making between humans and LLMs.
\end{enumerate}

The rest of the paper is organized as follows:

\begin{itemize}
\item Section~\ref{sec:relatedWork} reviews prior work on linguistic puzzle corpus creation and summarizes research on solving linguistic puzzles by both humans and large language models (LLMs).
\item Section~\ref{sec:corpusPairs} outlines the methodology we propose to convert Rosetta Stone puzzles into Match-Up format and details the resulting corpus of paired puzzles.
\item Section~\ref{sec:solving} presents experiments on solving the Rosetta Stone and Match-Up puzzles from Section~\ref{sec:corpusPairs}, comparing expert human and LLM performance on a shared subset and evaluating LLMs on additional pairs excluded from the human study.
\item Section~\ref{sec:evaluation} reports findings from follow-up interviews with the expert human solvers, documenting the strategies they employed to solve puzzles of different types.
\item Section~\ref{sec:conclusions} summarizes the contributions of this paper and concludes that, overall, a single generation procedure can be used to produce both Rosetta Stone and Match-Up puzzles, achieving varying levels of success across different linguistic topics.
\end{itemize}



\section{Related Work}
\label{sec:relatedWork}

\subsection{Existing Corpora}

The International Linguistics Olympiad (IOL), along with many national linguistic competitions, releases past puzzles together with their official solutions. Examples include NACLO (North America),\footnote{\url{https://naclo.org/practice.php}} \mbox{OzCLO} (Australia),\footnote{\url{https://ozclo.org.au/past-problems/}} UKLO (United Kingdom),\footnote{\url{https://www.uklo.org/past-exam-papers/}} etc.

The problems published by national competitions are incorporated into several datasets. For example, the \textsc{LingOly} dataset~\cite{LingOly} consists of puzzles originally created for UKLO. For each competition puzzle, UKLO provides solutions and a set of descriptive attributes, including puzzle difficulty (foundation, intermediate, advanced, etc.), linguistic topic (e.g., writing systems, morphology), question format (e.g., Rosetta Stone, Match-Up), language family, and other metadata. 

A recent work, \textsc{LingOly-TOO}~\cite{khouja2025lingolytoodisentanglingreasoningknowledge}, is an extension of the \textsc{LingOly} corpus. The \textsc{LingOly-TOO} corpus builds on the \textsc{LingOly} dataset by generating new puzzle variants with systematically introduced orthographic obfuscation.



An analysis of IOL puzzles~\cite{bozhanov-derzhanski-2013-rosetta} shows that the Rosetta Stone format is the most frequently used puzzle type. Moreover, IOL results indicate that “experienced solvers are better prepared to handle these [Rosetta Stone puzzles] than problems of other types.” 

The \textit{Puzz{L}ing Machines} dataset is a carefully curated resource consisting exclusively of \textbf{Rosetta Stone} puzzles from linguistic competitions for high school students across various countries~\cite{sahin-etal-2020-puzzling}.\footnote{\url{https://ukplab.github.io/PuzzLing-Machines/}}
 The dataset ``contains 96 unique puzzles from 81 languages that span 39 different language families from all over the world, as well as two creoles and two artificial languages.''

\textsc{modeLing} is another dataset that contains only \textbf{Rosetta Stone} puzzles~\cite{chi-etal-2024-modeling}. However, in contrast to the \textit{Puzz{L}ing Machines} dataset, \textsc{modeLing} includes newly created puzzles authored by expert puzzle writers. The main goal of the \textsc{modeLing} corpus is to create puzzles specifically for low-resource languages.


 \subsection{LLMs and Linguistic Puzzles}

{S}ahin et al.~\cite{sahin-etal-2020-puzzling} demonstrate that for the \textit{Puzz{L}ing Machines} dataset  ``both simple statistical algorithms and state-of-the-art deep neural models perform inadequately on this challenge''.

Over the past couple of years modern large language models (LLMs) have demonstrated impressive efficiency across a wide range of tasks~\cite{minaee2024large}. In text-related domains—such as understanding and analysis, generation and transformation, and conversational interaction—LLMs often outperform traditional pre-trained language models~\cite{zhou2024comprehensive}.

Linguistic puzzle datasets are increasingly being used as specialized benchmarks for evaluating the performance of large language models (LLMs). These puzzles are deliberately designed to be self-contained, ensuring that no external knowledge is required to solve them. This property makes linguistic puzzles an ideal testbed for assessing the reasoning capabilities of LLMs. Such evaluations are particularly informative when the puzzles involve low-resource languages~\cite{chi-etal-2024-modeling}.

The performance of LLMs on linguistic puzzle benchmarks depends heavily on the availability of resources for the language featured in the puzzles: ``the higher the resource level of the language, the better the scores''~\cite{LingOly}. LLM performance also depends on the linguistic topic at the core of the puzzle. Recent models often surpass human performance on topics such as compounding, number systems, morphology, phonology, semantics, and syntax. However, when puzzles are designed to test the ability to decipher rare or unfamiliar writing systems, humans still consistently outperform LLMs~\cite{EMNLP2025}.

In this paper, we use UKLO Rosetta Stone puzzles and convert them into corresponding Match-Up versions. We then evaluate both formats by comparing the performance of human solvers and large language models (LLMs) on each puzzle, examining how performance differs between the original Rosetta Stone puzzles and the generated synthetic Match-Up counterparts.

\section{Corpus}
\label{sec:corpusPairs}

In this project, we investigate whether Rosetta Stone and Match-Up puzzles represent genuinely distinct formats or simply reflect different perspectives on the same underlying puzzle structure. We focus on these two formats because they are the most frequently used in linguistic competitions. According to data released by UKLO, 45\% of all competition puzzles are of the Rosetta Stone type, while 28\% are Match-Up puzzles.

In our work, we treat each puzzle as a single unit. This approach differs from that adopted in the \mbox{\textsc{LingOly}} corpus, where each question within a puzzle is treated as an independent unit of analysis. For example, \textsc{LingOly} would consider the two translation questions shown in Table~\ref{tab:RosettaStoneExample} as separate items, whereas we evaluate overall performance across both questions within the same puzzle. Although it is possible to assess human and LLM performance on individual questions, participants in linguistic competitions cannot solve these questions in isolation. Instead, they must consider all examples (questions) within a puzzle collectively in order to infer the linguistic patterns necessary for solving the puzzle. Because one of the goals of this work is to determine if different puzzle formats require distinct generation procedures, it is essential to analyze each puzzle as a coherent whole.

\subsection{Conversion Procedure: from Rosetta Stone to Match-Up Format}
\label{sec:conversion}

We begin by collecting the Rosetta Stone puzzles from UKLO. In addition to the problem statements, we also compile the corresponding answers to the translation questions. For example, the information presented in Table~\ref{tab:RosettaStoneExample} is supplemented with the solutions to translation question \textbf{Q.5.3} (translate from Gilbertese to English). Specifically, 
   \begin{enumerate}
   \setcounter{enumi}{13}
     \item \textit{Women will play tomorrow}\\ Solution: \textbf{A takaakaro aiine ningaabong}
     \item \textit{You are sitting next to the store today}\\ Solution: \textbf{Ko tekateka irarikin te titooa \ng{}koe n te bong aei}
    \end{enumerate}

In total, the final version of the puzzle consists of 12 pairs of Gilbertese sentences and their corresponding English translations. We then randomly shuffle the English sentences, producing a set of 12 Gilbertese sentences and a corresponding set of 12 shuffled English sentences labeled A through L. This transformation results in a Match-Up puzzle, where the task is to match each Gilbertese sentence with its correct English translation.

In addition to the aligned text pairs used in Rosetta Stone puzzles, each UKLO puzzle includes a short contextual description (preamble), a brief informational note about the language under analysis. This preamble typically mentions the language family of the target language or highlights particular linguistic features that may or may not be required for solving the puzzle. When generating a Match-Up puzzle corresponding to a Rosetta Stone puzzle, we preserve this preamble information. Consequently, the final version of the Match-Up puzzle corresponding to the 2023 Gilbertese puzzle includes the following preamble:

\begin{quote}{The Gilbertese language is an Austronesian language spoken in Kiribati, a country consisting of a number of islands lying to the northeast of Australia.}
\end{quote}

Thus, the Match-Up Gilbertese puzzle that corresponds to the Rosetta Stone puzzle presented in Table~\ref{tab:RosettaStoneExample} is presented in Table~\ref{tab:RosettaToMatchUpExample}.

\begin{table*}[t!]
\centering
\begin{tabularx}{\textwidth}{llll} 
\toprule
\multicolumn{4}{p{6in}}{The Gilbertese language is an Austronesian language spoken in Kiribati, a country consisting of a number of islands lying to the northeast of Australia. Below are some sentences in Gilbertese, followed by their English translations in a random order. }\\
 \hline

 \multicolumn{1}{c}{\textbf{ }} & 
\multicolumn{1}{c}{\textbf{Gilbertese}} &
\multicolumn{1}{c}{ } &
\multicolumn{1}{c}{\textbf{English}} \\
 
  \hline

1  & \textbf{Ko nakonako \ng{}koe} & A & \textit{Women will play 
tomorrow} \\ 
  \hline

2  & \textbf{E nakonako te aiine}   & B & \textit{You are walking} \\ 
 \hline

3  & \textbf{I takaakaro \ng{}ai }  & C & \textit{A woman is walking} \\ 
 \hline

4  & \textbf{E nakonako nakon te titooa Meeri} & D  & \textit{People are bathing today} \\ 
  \hline

5  & \textbf{A tekateka irarikin te auti aiine}  & E & \textit{You are sitting next to the 
store today} \\ 
  \hline

6  & \textbf{A tebotebo nakekei n te bong aei}   & F & \textit{You will sit tomorrow} \\
 \hline

7  & \textbf{I tebotebo inanon te auti \ng{}ai}  & G & \textit{Mary is walking to the 
store} \\ 
 \hline

8  & \textbf{A takaakaro inanon te titooa ataei}   & H & \textit{I am bathing in the house} \\ 
 \hline

9  & \textbf{Ko tekateka \ng{}koe ningaabong}   & I & \textit{Children are playing in the 
store} \\ 
  \hline

10  & \textbf{E takaakaro irarikin te kawai te ataei }  & J & \textit{Women are sitting next to 
the house} \\ 
      & \textbf{n te bong aei}   & & \textit{ } \\
 \hline

11  & \textbf{A  takaakaro aiine ningaabong}   & K & \textit{The child is playing next to 
the road today} \\ 
      & \textbf{n te bong aei}   & & \textit{ } \\ 
 \hline

12  & \textbf{Ko tekateka irarikin te titooa \ng{}koe}  & L & \textit{I am playing} \\ 
      & \textbf{n te bong aei.}  & & \textit{ } \\ 
  \hline

\multicolumn{4}{p{6in}}{\textbf{Q.1} Determine the correct correspondence. (A to L) }\\
\\
\multicolumn{4}{c}{
\begin{tabular}{|l|l|l|l|l|l|l|l|l|l|l|l|l|}
        \hline
        Gilbertese & A & B & C & D & E & F & G & H & I & J & K & L\\
        \hline
        English &  &   &    &   &   & & & & & & &   \\
        \hline
\end{tabular}
} \\
\\
\bottomrule

\end{tabularx}
\caption{Match-Up Pair for the 2023 UKLO Gilbertese Rosetta Stone puzzle.}
\label{tab:RosettaToMatchUpExample}
\end{table*}

The solution to the Match-Up puzzle is a table with matched pairs. For example, the answer for the Polish puzzle from Table~\ref{tab:MatchUpExample} is the following:\footnote{\url{https://www.uklo.org/wp-content/uploads/2022/05/2015_3.-Polish.pdf}} 
\begin{center}
\begin{tabular}{|l|l|l|l|l|l|l|}
        \hline
        Polish & A & B & C & D & E & F \\
        \hline
        English & 4  & 6   & 2   & 5  & 1  & 3 \\
        \hline
\end{tabular}
\end{center}

It must be noted that not all puzzles contain complete sentences. For example, the Rosetta Stone Permyak puzzle used in the 2023 UKLO\footnote{\url{https://www.uklo.org/wp-content/uploads/2023/03/2023_R1_5-Permyak.pdf}} consists of pairs of strings shorter than full sentences. Here are the Permyak--English text string pairs:\\

\begin{tabular}{ l  l}
\hline
 \textbf{Permyak} & \textit{English}  \\ 
 \hline
  \textbf{k’erkulan’}            & \textit{towards the house }              \\
 \textbf{p\st{i}zannez\st{i}tl\textschwa{}n}  & \textit{of your (sg.) desks }      \\
 \textbf{pon\st{i}t}        & \textit{your (sg.) dog }              \\
 \textbf{purtn\st{i}s}        & \textit{their knife }             \\
\textbf{k\textschwa{}innez\st{i}s }        & \textit{his wolves  }             \\
\textbf{v\textschwa{}r\textschwa{}l\textschwa{}n}        & \textit{of my forest  }             \\
\textbf{purt\textschwa{}la}        & \textit{for the sake of my knife  }             \\
\textbf{t\st{i}ez\st{i}tk\textschwa{}t}    & \textit{with your (sg.) lakes  }             \\
\textbf{k’erkuezlis’}        & \textit{from the houses  }             \\
\textbf{jus’s’ez\textschwa{}}        & \textit{ my swans }             \\
\textbf{kok\st{i}sk\textschwa{}t}        & \textit{with his foot  }             \\
\textbf{k’i\st{i}tlan’ }        & \textit{towards your (sg.) hand  }             \\
 \hline%
 \label{tab:permyak}
\end{tabular}

\subsection{Corpus Description}

To construct our corpus, we collected 96 Rosetta Stone puzzles published on the UKLO website as of October 2025, excluding those from 2010 and 2011 due to the absence of participant performance data. From this set, 30 puzzles (with their corresponding Match-Up conversions) were selected for the experiments, based on the design of the human evaluation study (see Section~\ref{sec:humanEvaluationExperiment}). The experimental subset is balanced with respect to puzzle difficulty and linguistic topic. 

The full dataset contains 192 puzzle files, corresponding to 96 Rosetta Stone puzzles and their Match-Up counterparts. Each entry includes a preamble, the puzzle questions, and reference answers. In both the human and LLM experiments, only the preamble and questions are provided to participants, while the reference answers are used solely for evaluation. We rely on the puzzle format labels (e.g., Rosetta Stone) provided by the original puzzle authors. However, not all UKLO puzzles strictly follow the canonical Rosetta Stone format discussed in this paper. As a result, several puzzles cannot be converted into Match-Up format. Any deviations or special structural features are explicitly documented.

\section{Solving Rosetta Stone and Match-Up Puzzle Pairs}
\label{sec:solving}

Using the generated corpus (Section~\ref{sec:corpusPairs}), we evaluate how converting Rosetta Stone puzzles into the Match-Up format affects human and LLM performance on puzzle-solving tasks.

\subsection{Human Evaluation Experiment}
\label{sec:humanEvaluationExperiment}

For the human evaluation, we engaged two accomplished Linguistic Olympiad participants with \mbox{NACLO} experience but no prior exposure to UKLO puzzles.\footnote{It must be pointed out that there is a small set of puzzles that were used in both NACLO and UKLO.} The evaluators were not informed that the Match-Up puzzles were derived from the corresponding Rosetta Stone versions and were simply asked to solve a set of linguistic puzzles. 

The number of human annotators was limited due to the specialized expertise required for the task. Linguistic puzzle solving, particularly across multiple languages, linguistic domains, and difficulty levels, demands advanced analytical skills and prior experience with competition-style puzzles (e.g., foundation through advanced levels). Consequently, annotators were selected based on demonstrated proficiency in solving such puzzles rather than general linguistic background alone.

Our evaluation protocol did not aim to measure inter-annotator agreement over multiple competing solutions, as is common in labeling tasks. Instead, the goal was to assess \textit{solvability}. Specifically, we evaluated whether synthetic Match-Up puzzles, automatically converted from Rosetta Stone puzzles, could be successfully solved by qualified human solvers. The outcome measure was binary at the puzzle level: a puzzle was considered solvable if at least one expert solver arrived at a correct and complete solution using only the information provided in the puzzle. Under this framework, difficulty is orthogonal to validity. While some puzzles may require substantial reasoning effort, the existence of a correct solution obtained independently by a qualified solver provides evidence that the generated puzzle is well-formed and solvable.

This evaluation design aligns with the primary objective of the study: to determine whether the conversion procedure preserves logical structure and inferential sufficiency, rather than to assess solution variability across individuals.

The experiment consisted of two stages distinguished by puzzle difficulty. UKLO defines five difficulty levels: \textit{Breakthrough}, \textit{Foundation}, \textit{Intermediate}, \textit{Advanced}, and \textit{Round~2}, with some puzzles spanning adjacent levels. Stage 1 used easier puzzles labeled \textit{Breakthrough/Foundation} or \textit{Foundation/Intermediate}, while Stage 2 used more complex \textit{Advanced} or \textit{Round~2} puzzles.

To ensure that all parameters other than difficulty remain constant, both stages include puzzles centered on the same linguistic topic or combination of topics. Each stage includes two puzzles per linguistic topic (or topic combination). The lists below present the linguistic topics used in the human evaluation experiments and the number of puzzles selected for each stage.


\begin{center}
\begin{tabular}{ l  c c }
\hline
\hline
 \textbf{Topic} & \textit{Stage 1}  & \textit{Stage 2} \\ 
 \hline
Morphology & 2  & 2  \\
\hline
Syntax  & 2  & 2  \\
\hline
Syntax and Morphology   & 2  & 2  \\
\hline
Syntax and Semantics    & 2  & 0  \\
\hline
Semantics, Morphology,    & 0  & 2  \\
and Syntax    & &  \\
 \hline%
\end{tabular}
\end{center}

As shown in the table above, for the  \textbf{syntax and semantics} combination in Stage 1, we were unable to identify puzzles with the exact same topic combination for Stage 2. Therefore, in Stage 2 (the more difficult set), instead of using puzzles focused solely on \textbf{syntax and semantics}, we include puzzles that combine \textbf{morphology, syntax, and semantics}.

The study includes 16 puzzle pairs (32 puzzles), evenly split across two stages. Evaluators never see both puzzles from the same pair; for each topic per stage, they solve one Rosetta Stone and one Match-Up puzzle generated from different source puzzles on the same topic.

Table~\ref{tab:combined_ros_scores} shows human and LLM performance on the 16 Rosetta Stone/Match-Up puzzle pairs, reported as the percentage of correctly answered questions per puzzle following UKLO’s convention. Details of the LLM evaluation appear in Section~\ref{sec:LLMEvaluationExperiment}.


Table~\ref{tab:combined_ros_scores} compares average UKLO performance (\textbf{UKLO}) with that of the two human evaluators on the same puzzles. For dual-level puzzles (e.g., Breakthrough/Foundation), the higher UKLO score is used. Because some UKLO puzzles, such as the 2024 Coptic puzzle,\footnote{\url{https://www.uklo.org/wp-content/uploads/2024/03/2024-R2_4-Coptic.pdf}} include additional questions, UKLO’s aggregate scores may reflect tasks beyond the Rosetta Stone format. Columns \textbf{HE (RS)} and \textbf{HE (MU)} report evaluator accuracy on the original and converted puzzles, respectively.

As expected, average performance for both UKLO participants and our evaluators declines from Stage~1 to Stage~2 due to the higher difficulty of the Stage~2 puzzles.

The results in Table~\ref{tab:combined_ros_scores} show that overall human performance remains stable when solving the synthetic Match-Up puzzles, with no decline relative to the original Rosetta Stone puzzles. Notably, we use a strict scoring procedure for the Rosetta Stone puzzles, counting only answers that exactly match the official UKLO key. Interestingly, the Match-Up puzzles display an all-or-nothing performance pattern among human evaluators.

\begin{table*}[t]
\centering
\begin{tabular}{llcccc||ccc}
\hline

\textbf{Topic} & \textbf{Stage} & \textbf{UKLO} & \textbf{GPT5} & \textbf{Gem2.5-} & \textbf{HE} & \textbf{GPT5} & \textbf{Gem2.5-} & \textbf{HE} \\
\textbf{} & \textbf{} & \textbf{} & \textbf{} & \textbf{pro} & \textbf{(RS)} & \textbf{} & \textbf{pro} & \textbf{(MU)} \\

\hline
\multicolumn{2}{c}{} & \multicolumn{4}{c||}{\textbf{Rosetta Stone}}  &  \multicolumn{3}{c} {\textbf{Match-Up}} \\
\multicolumn{2}{c}{} & \multicolumn{4}{c||}{\textbf{Original Puzzle}}  &  \multicolumn{3}{c} {\textbf{Conversion}} \\
\hline
\hline

\multirow{2}{*}{Morphology} & s1 & 76.5 & 97.5 & 100 & 100 & 100 & 100 & 96 \\
\cmidrule(l){2-9} 
 & s2 & 39.5 & 77.5 & 93 & 97.5 & 2 & 0 & 50 \\
\hline
{Syntax,} & s1 & 47 & 81.5 & 85.5 & 94 & 100 & 100 & 100 \\
\cmidrule(l){2-9}
 {Morphology}& s2 & 21 & 71 & 87 & 97 & 22 & 96 & 100 \\
\hline
{(Morphology), Syntax } & s1 & 54.5 & 78.5 & 84.5 & 87.5 & 18.5 & 16 & 100 \\
\cmidrule(l){2-9}
 {Semantics} & s2 & 37.5 & 67.5 & 72 & 91 & 4 & 100 & 50 \\
\hline
\multirow{2}{*}{Syntax} & s1 & 64 & 90 & 81.5 & 81.5 & 89.5 & 89.5 & 100 \\
\cmidrule(l){2-9}
 & s2 & 52 & 64 & 72 & 89 & 56.6 & 63 & 100\\
\hline

\end{tabular}
\caption{Average Scores by Linguistic Topic and Stage on 16 Pairs of Puzzles. 
UKLO - The average human performance reported on the UKLO website; GPT5 and Gem2.5-pro - The average performance by GPT5 and Gemini2.5-pro respectively; HE (RS) - The average performance of the project human evaluators on the UKLO Rosetta Stone puzzles; HE (MU) - The average performance of the project human evaluators on the synthetic Match-Up puzzles.}
\label{tab:combined_ros_scores}
\end{table*}

\subsection{Large Language Models Experiments}
\label{sec:LLMEvaluationExperiment}

We evaluate two state-of-the-art LLMs—OpenAI’s GPT-5~\cite{gpt5} and Google’s Gemini~2.5-Pro~\cite{comanici2025gemini}, under identical zero-shot settings. For Rosetta Stone puzzles, models receive a preamble and translation strings; for Match-Up puzzles, they are given a preamble and two sets of strings to match.

\subsubsection{Experiment 1}

Table~\ref{tab:combined_ros_scores} reports the average performance across two puzzles at each stage for each linguistic topic (or topic combination) for OpenAI’s GPT-5 (column GPT-5) and Google’s Gemini 2.5-Pro (column Gem 2.5-Pro). These are the two SOTA publicly available LLMs at the time of the experiments (October 2025). When evaluating Match-Up puzzles, we follow the strict evaluation procedure described in~\cite{EMNLP2025}: if a model’s output is presented in a perfectly alphabetical order, we assign a score of 0 to such responses, even if some matches are accidentally correct.

For both LLMs, performance on the Rosetta Stone puzzles, similar to that of human participants, generally decreases from Stage~1 to Stage~2. Overall, Gemini 2.5-Pro outperforms GPT-5 across all linguistic topics except one, namely Syntax. However, for the synthetic Match-Up puzzles, the results across the two stages and linguistic topics are more variable and warrant further investigation.

The two most noteworthy cases are those involving the Morphology topic and the combination of Semantics, Morphology, and Syntax topics. For Morphology, both models achieve 100\% accuracy on Stage~1; however, their performance on Stage~2 drops dramatically to 2\% and 0\%, respectively. For the Semantics/Morphology/Syntax combination, both models perform relatively poorly on Stage~1 (18.5\% and 16\%), while on Stage~2, GPT-5’s accuracy falls further to 4\%, whereas Gemini~2.5-Pro’s accuracy increases sharply to 100\%. Following~\cite{LingOly}, we assume that these contrasting results are influenced by the specific set of languages used in the puzzles for this combination of linguistic topics.

\begin{table*}[t]
\centering
\begin{tabular}{lcccc||cc}
\hline
\textbf{Topic} & \textbf{Difficulty} & \textbf{UKLO} & \textbf{GPT5} & \textbf{Gem2.5-pro} & \textbf{GPT5} & \textbf{Gem2.5-pro}  \\

\hline

\multicolumn{2}{c}{} & \multicolumn{3}{c||}{\textbf{Rosetta Stone Original Puzzle}}  &  \multicolumn{2}{c} {\textbf{Match-Up Conversion}} \\
\hline
\hline

{}& s1 & 61 & 84.6 & 84.6 & 90 & 15 \\
\cmidrule(l){2-7}
{} & s2 & 16 & 98.1 & 96.3 & 100 & 100 \\
\cmidrule(l){2-7}
{Syntax, Morphology}&  s2 & 80 & 100 & 87.9 & 100 & 100 \\
\cmidrule(l){2-7} 
\multirow{2}{*}{}& s2 & 32 & 56.4 & 64.1 & 0 & 0 \\
\cmidrule(l){2-7}
{}& s2 & 12 & 62.5 & 64.6 & 0 & 100 \\
\hline 

\multirow{4}{*}{Syntax} & s1 & 45 & 100 & 100 & 100 & 100  \\
\cmidrule(l){2-7}
 & s1 & 70 & 87.5 & 91.7 &  100 & 100\\
 \cmidrule(l){2-7}
 & s2 & 96 & 100 & 95.8 &  100 & 100\\ 
 \cmidrule(l){2-7}
 & s2 & 46 & 90.0 & 100 &  30 & 100\\

 \hline

 \multirow{3}{*}{Morphology}& s1 & 44 & 62.5 & 95.8 &  0 & 0\\ 
  \cmidrule(l){2-7}
& s2 & 27 & 93.7 & 90.6 &  29.4 & 0\\  
\cmidrule(l){2-7}
& s2 & 31 & 93.3 & 68.9 &  100 & 100\\  
 \hline

{Morphology, Semantics} & s2 & 24 & 70.8 & 56.2 &  5.3 & 5.3\\  
 \hline

{Phonology} & s2 & 60 & 93.3 & 68.9 &  100 & 100\\  
\hline
\end{tabular}
\caption{Scores by Linguistic Topic and Stage on 14 Pairs of Puzzles. 
UKLO - The human performance reported on the UKLO website; GPT5 and Gem2.5-pro - performance by Chat-GPT5 and Gemini~2.5-pro respectively.}
\label{tab:additional_14_puzzles}
\end{table*}
\subsubsection{Experiment 2}

In addition to the 16 puzzle pairs evaluated in Table~\ref{tab:combined_ros_scores}, we apply OpenAI’s GPT-5 and Google’s Gemini~2.5-Pro to an additional set of 14 pairs of original UKLO Rosetta Stone puzzles and corresponding synthetic Match-Up puzzles. The results of running OpenAI’s GPT-5 and Google’s Gemini~2.5-Pro on the additional set of 14 pairs of puzzles are displayed in Table~\ref{tab:additional_14_puzzles}. Due to the nature of the UKLO dataset, it is impossible to create an additional balanced set of linguistic puzzles across the difficulty and the linguistic topic dimensions. Moreover, not all Rosetta Stone puzzles can be easily translated into the corresponding Match-Up puzzles. For example, the 2014 Kairak problem\footnote{\url{https://www.uklo.org/wp-content/uploads/2022/08/2014.6-Kairak.pdf}} has three types of verbal patterns translated into English and cannot be converted into a Match-Up format following the procedure described in Section~\ref{sec:corpusPairs}.

Table~\ref{tab:additional_14_puzzles} reports results for 14 pairs of UKLO Rosetta Stone puzzles and their corresponding Match-Up versions, organized by linguistic topic and stage (Stage~1 or Stage~2; see Section~\ref{sec:humanEvaluationExperiment}). For puzzles assigned to two difficulty levels, the higher UKLO score is used. The table includes GPT-5 and Gemini~2.5-Pro performance on both puzzle types. As in Table~\ref{tab:combined_ros_scores}, LLMs outperform UKLO participants on Rosetta Stone puzzles and show comparable results to each other. For Match-Up puzzles, both models exhibit an all-or-nothing pattern, particularly on morphology-related puzzles. We believe it might be due to the shorter text strings typical of this topic (see the 2013 Permyak puzzle example in Section~\ref{sec:conversion}). The all-or-nothing trait exhibited by both humans and LLMs requires further investigation.

\section{Approaches Towards Solving Linguistic Puzzles}
\label{sec:evaluation}

After completing the experiment described in Section~\ref{sec:humanEvaluationExperiment}, the human evaluators were interviewed about their experience with puzzles of different formats and the strategies they used to solve them. At this stage, the evaluators were not aware that the Match-Up puzzles had been derived from the corresponding Rosetta Stone puzzles. Both evaluators confirmed that the synthetic Match-Up puzzles appeared to be plausible linguistic puzzles. 

The approaches used to solving puzzles varied depending on the puzzle format and the linguistic topic involved. Below we summarize several strategies commonly employed by the evaluators when solving Match-Up puzzles.

\textbf{Character and word count} While languages may encode the same meaning using strings of different lengths, human solvers often rely on the heuristic that longer strings correspond across languages. This approach can be helpful but not always reliable. Match-Up puzzles with few tokens (e.g., the UKLO 2013 Permyak puzzle in Section~\ref{sec:conversion}) are especially difficult and reward familiarity with the language or its family. Such puzzles frequently center on morphology, the category that notably produces the ``all-or-nothing'' performance pattern observed in both humans and LLMs.

\textbf{Repeating Matching Strings} Another effective strategy involves counting occurrences of identical substrings that are likely to correspond to the same English words in translation.

\textbf{Use of proper names} Proper names frequently serve as anchors for identifying correspondences between languages. For instance, in Table~\ref{tab:RosettaStoneExample}, English \textit{Mary} corresponds to Gilbertese \textit{Meeri}; in Table~\ref{tab:MatchUpExample}, English \textit{Alice} corresponds to Polish \textit{Alicja}, and \textit{Peter} corresponds to \textit{Piotr}. Proper names thus often provide initial clues in both Rosetta Stone and Match-Up puzzles, an observation that should be taken into account when designing automatic methods for linguistic puzzle generation.

\section{Conclusion}
\label{sec:conclusions}

We test the hypothesis that two linguistic puzzle formats, Rosetta Stone and Match-Up, represent complementary views of the same underlying structure. To investigate this, we develop a systematic conversion procedure and apply it to create a corpus of 96 paired puzzles, each consisting of an original Rosetta Stone puzzle and its Match-Up counterpart.

We evaluate the quality of the generated Match-Up puzzles through a series of experiments on 30 puzzles. Sixteen of the 30 puzzle pairs are solved by human evaluators, and all 30 are solved by large language models (LLMs). Based on the experimental results and follow-up interviews with the evaluators, we conclude: the proposed conversion procedure can effectively generate Match-Up puzzles from existing Rosetta Stone puzzles. However, further research is required to assess its applicability across a broader range of linguistic topics.

The experiments reveal an all-or-nothing performance pattern in both humans and LLMs, particularly on morphology-focused puzzles. This pattern suggests that Match-Up puzzles capture a distinct type of linguistic reasoning. Together, the dataset and methodology introduced here provide a foundation for automated puzzle generation and for future studies comparing human and machine problem-solving across languages and puzzle formats.

\section{Ethics Statement}
This study involved two experienced high‑school lingusitic puzzle solvers. Written informed consent (and parental/guardian consent for minors) was obtained; no personal data were collected or stored. Tasks used publicly available competition puzzles. No sensitive attributes were elicited. The released dataset contains only problem statements, preambles, and answer keys from public sources, with metadata derived from UKLO and converted formats; it contains no personal data.

\section{Limitations}
Human study (N=2) limits inference; language/topic coverage is imbalanced; some Rosetta puzzles (e.g., multi‑template verb systems) cannot be converted without additional heuristics; LLM scores may vary across versions and decoding settings; and our strict evaluation for Rosetta (exact‑match only) may undercount partial progress.

\section{Data and Code Availability}

The data is realized in:\\ {\footnotesize\url{https://github.com/ef2020/lrec2026-data}}

\nocite{*}
\section{Bibliographical References}\label{sec:reference}

\bibliographystyle{lrec2026-natbib}
\bibliography{lrec2026-example}


\end{document}